\documentclass[10pt,twocolumn,letterpaper]{article}

\usepackage{iccv}
\usepackage{times}
\usepackage{epsfig}
\usepackage{graphicx}
\usepackage{amsmath}
\usepackage{amssymb}
\usepackage[ruled,linesnumbered]{algorithm2e}

\usepackage{mathrsfs}
\usepackage{booktabs}
\usepackage{algpseudocode}
\usepackage{multirow}
\usepackage{caption}
\usepackage{subcaption}
\usepackage{diagbox}
\usepackage{bm}
\usepackage{color}

\usepackage{amsfonts}
\usepackage{capt-of,etoolbox}

\usepackage[pagebackref=true,breaklinks=true,letterpaper=true,colorlinks,bookmarks=false]{hyperref}

\iccvfinalcopy 

\def\iccvPaperID{3491} 

\ificcvfinal\fi

\makeatletter
\apptocmd\@maketitle{{\myfigure{}\par}}{}{}
\makeatother

\begin{document}

\newcommand\myfigure{
\centering
\includegraphics[width=1.0\linewidth]{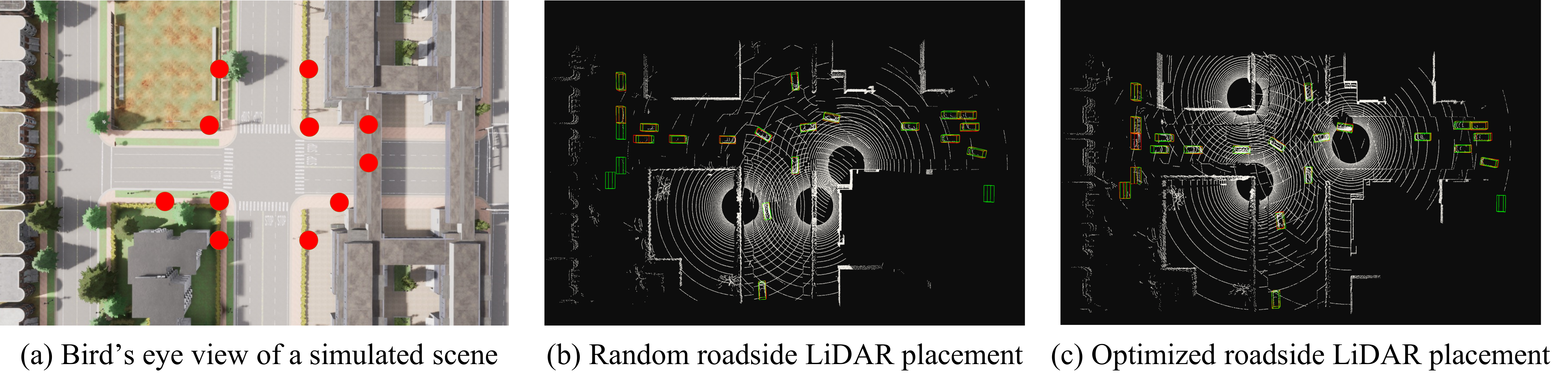}
\vspace{-3mm}
\captionof{figure}{
    An example of roadside LiDAR placement optimization. 
    (a) is a simulated scene in CARLA, where red circles indicate potential LiDAR deployment locations.
    We perform a random selection as well as our proposed method to select $3$ locations for LiDAR setup, as shown in (b) and (c). The \textcolor{green}{green} and \textcolor{red}{red} bounding boxes denote the ground truth vehicles and the detected vehicles using the detection model, respectively.
    It can be observed that by optimizing the placement, the red box overlaps more with the green box, indicating an improvement in the detection performance.
}
\label{fig1}
\vspace{6mm}
}

\title{Optimizing the Placement of Roadside LiDARs for Autonomous Driving}


\author{
Wentao Jiang\textsuperscript{1,2,4}
\ Hao Xiang\textsuperscript{\rm 3}
\ Xinyu Cai\textsuperscript{\rm 2}
\ Runsheng Xu\textsuperscript{\rm 3}
\\
\ Jiaqi Ma\textsuperscript{\rm 3}
\ Yikang Li\textsuperscript{\rm 2}\thanks{Equal corresponding author}
\ Gim Hee Lee\textsuperscript{\rm 4}
\ Si Liu\textsuperscript{\rm 1}\footnotemark[1]
\\
\textsuperscript{\rm 1} Beihang University \quad
\textsuperscript{\rm 2} Shanghai AI Laboratory \quad
\textsuperscript{\rm 3} UCLA  \quad
\textsuperscript{\rm 4} NUS
}

\maketitle
\ificcvfinal\thispagestyle{empty}\fi

\vspace{-4mm}
\begin{abstract}
    Multi-agent cooperative perception is an increasingly popular topic in the field of autonomous driving, where roadside LiDARs play an essential role. However, how to optimize the placement of roadside LiDARs is a crucial but often overlooked problem.  This paper proposes an approach to optimize the placement of roadside LiDARs by selecting optimized positions within the scene for better perception performance.
    To efficiently obtain the best combination of locations, a greedy algorithm based on  \textit{perceptual gain} is proposed, which selects the location that can maximize the perceptual gain sequentially.
    We define perceptual gain as the increased perceptual capability when a new LiDAR is placed.
    To obtain the perception capability, we propose a \textit{perception predictor} that learns to evaluate LiDAR placement using only a single point cloud frame. A dataset named Roadside-Opt is created using the CARLA simulator to facilitate research on the roadside LiDAR placement problem.
    Extensive experiments are conducted to demonstrate the effectiveness of our proposed method.
\end{abstract}


\section{Introduction}

Accurate perception of the driving environment is critical for autonomous driving, and recent advancements in deep learning have significantly improved the robustness of single-vehicle perception algorithms in tasks such as object detection \cite{lang2019pointpillars,2018Complex,2017Vote3Deep,2018YOLO3D,2017PointNet,2014Beyond,2017VoxelNet}. Compared to single-vehicle perception, utilizing both vehicle and infrastructure sensors for collaborative perception brings significant advantages, such as a global perspective beyond the current horizon and coverage of blind spots.
Current progress in multi-agent communication has enabled such collaboration by sharing the visual data with other agents \cite{xu2022v2x,bhover2017v2x,olaverri2018connection,xiang2022v2xp}. Among the collaboration, the LiDAR sensors play a critical role as they can provide reliable geometry information and be directly transferred to real-world coordinate systems from any agent's perspective\cite{9548787}.

To optimize the LiDAR perception performance, the placement of LiDAR sensors is critical \cite{hu2022investigating}. However, current research mainly focuses on the placement of LiDAR sensors on vehicles and neglects the placement problem for roadside LiDAR \cite{hu2022investigating,ma2021perception}. As Vehicle-to-Everything (V2X) applications continue to develop, it becomes increasingly important to determine the optimal placement for roadside LiDARs to maximize their benefits. Unlike vehicle LiDAR, the placement of roadside LiDAR has greater flexibility. As depicted in Figure~\ref{fig1}, the roadside LiDAR placement optimization problem aims to maximize perception ability for a scene, usually an intersection with dense traffic flow, by selecting $M$ optimal locations from $N$ potential locations for placing roadside LiDARs.

Existing literature on LiDAR placement such as \cite{hu2022investigating,ma2021perception} propose surrogate metrics such as S-MIG and Perception Entropy to evaluate LiDAR placement, while \cite{cai2022analyzing} analyzes the correlation between perception performance and LiDAR point cloud density and uniformity. However, these methods cannot be directly applied to roadside LiDAR optimization since they only evaluate LiDAR placement rather than directly optimizing sensor position. Additionally, these surrogate metrics are based on probabilistic occupancy and may not be directly related to Average Precision (AP) obtained by multi-agent detection models, which may result in sub-optimal performance when tested with multi-agent 3D detection models.

To address these challenges, we propose a novel method to directly optimize the placement of roadside LiDAR through a \textit{perceptual gain} based greedy algorithm and a \textit{perception predictor} that learns to evaluate placements quickly.
To optimize the placement of roadside LiDAR, it is necessary to tackle the efficiency problem.
A straightforward method is to use the brute-force algorithm, which searches all possible combinations of selected locations (\ie, $C^M_N$ choices in total).
However, with the number of selected LiDAR $M$ and potential setup locations $N$ increasing, the runtime of the brute-force algorithm becomes unacceptable.
Fortunately, we observe that if we select the LiDAR placement sequentially, the obtained perceptual performance exhibits an intuitive diminishing returns property, also known as submodularity \cite{krause2014submodular,roberts2017submodular}.

Our proposed \textit{perceptual gain} based greedy algorithm uses this property to obtain approximate optimal solutions that dramatically reduce computation time. 
Specifically, we place the LiDAR sequentially and choose the potential location to maximize the perceptual gain.
The perceptual gain is defined as the increase in perception ability when a newly added LiDAR is placed. 
To obtain perceptual gain, testing the LiDAR configuration on datasets and calculating Average Precision (AP) is a straightforward way.
However, it is still inefficient because calculating the AP on a large dataset is also time-consuming.
To overcome this challenge, we develop a novel \textit{perception predictor} that can predict the perception ability of a LiDAR placement using only a single frame of point cloud data. 
Additionally, we create a dataset called ``Roadside-3D'' using the realistic self-driving simulator, CARLA~\cite{dosovitskiy2017carla} since no suitable dataset for roadside LiDAR optimization research exists.
We conduct extensive experiments to demonstrate the efficiency and effectiveness of our proposed method.

Our contributions are summarized as follows:
\begin{itemize}
    \item To the best of our knowledge, we are the first to study the optimization problem of roadside LiDAR placement.
    We build the ``Roadside-3D'' dataset using the CARLA simulator to facilitate the related research.
    \item We propose a \textit{perceptual gain} based greedy algorithm that obtains approximate optimal solutions for roadside LiDAR placement optimization. This approach dramatically reduces the computational time compared to the simple brute-force algorithm.
    \item We introduce a novel \textit{perception predictor} that can quickly obtain the perceptual gain by predicting the perception ability of a LiDAR placement.
\end{itemize}

\section{Related Work}

\textbf{Multi-agent/V2X perception.}
Multi-agent/V2X perception aims at detecting objects by leveraging the sensing information from multiple connected agents.
Current research on V2X perception mainly focuses on V2V (vehicle-to-vehicle) and V2I (vehicle-to-infrastructure).
Multi-agent perception can be broadly categorized into early, intermediate, and late fusion.
Early fusion \cite{gao2018object} directly transforms the raw data and fuses them to form a holistic view for cooperative perception.
Intermediate fusion \cite{li2021learning,chen2017multi,wang2020v2vnet,xu2022opv2v,xu2022v2x,lei2022latency,Su2022uncertainty} extract the intermediate neural features from each agent's observation and then broadcast these features for feature aggregation while late fusion \cite{melotti2020multimodal, fu2020depth, caltagirone2019lidar, 2012Car2X} circulates the detection outputs and uses non-maximum suppression to aggregate the received predictions to produce final results. Although early fusion can achieve outstanding~\cite{wang2020v2vnet} performance as it can preserve complete raw information, it requires large transmission bandwidth to transmit raw LiDAR point clouds, which makes it infeasible to deploy in the real world.
On the other hand, late fusion only requires small bandwidth, but its performance is constrained due to the loss of environmental context. Therefore, intermediate fusion has attracted increasing attention due to its good balance between accuracy and transmission bandwidth.
V2VNet \cite{wang2020v2vnet} proposes a spatially aware message-passing mechanism to jointly reason detection and prediction.
\cite{2020Learning} regresses vehicle localization errors with consistent pose constraints to mitigate outlier messages.
DiscoNet \cite{li2021learning} uses knowledge distillation to improve the intermediate process. 
V2X-ViT \cite{xu2022v2x} presented a unified transformer architecture for V2X perception that can capture the heterogeneous nature of V2X systems with strong robustness against various noises.
This paper uses early, intermediate, and late fusion to evaluate the average precision of selected roadside LiDAR placement.

\textbf{LiDAR placement.}
Previous research on LiDAR placement has mainly focused on analyzing and evaluating the placement of vehicle LiDARs for autonomous driving.
Ma \etal \cite{ma2021perception} proposed a new method based on Bayesian theory conditional entropy to evaluate the vehicle sensor configuration.
Liu \etal \cite{liu2019should} proposed a bionic volume to surface ratio (VSR) index to solve the optimal LiDAR configuration problem of maximizing utility.
Dybedal \etal \cite{dybedal2017optimal} used the mixed integer linear programming method to solve the problem of finding the best placement position of 3D sensors.
Hu \etal \cite{hu2022investigating} proposed a method for optimizing the placement of vehicle-mounted LiDAR and compared the perceptual performance effects of many traditional schemes.
Kim \etal \cite{kim2019placement} investigated how to increase the number of point clouds in a given area and reduce the dead zone as much as possible.
Cai \etal \cite{cai2022analyzing} propose a LiDAR simulation library for the simulation of infrastructure LiDAR sensors. They also analyze the correlation between perception performance and the density/uniformity of the LiDAR point cloud.
However, it is not feasible to adapt the existing works to our roadside LiDAR optimization problem since they only propose methods to evaluate different LiDAR placements instead of directly optimizing and searching sensor positions.
To this end, we present a novel efficient greedy algorithm to search the optimal positions for roadside LiDAR.

\section{Method}

To optimize roadside LiDAR placement, we aim to place $M$ LiDAR sensors at $N$ potential locations where roadside LiDAR can be deployed (\eg, street lights, traffic signals, and traffic signs in the scene) such that the optimized placement can maximize the perceptual performance of downstream multi-agent 3D detection models.
To avoid using the brute-force algorithm to evaluate every combination of possible locations, we design a novel perceptual gain based greedy algorithm that dramatically reduces computation time.
Specifically, we greedily select the location that can maximize the perceptual gain sequentially.
The perceptual gain is defined as the increase in perception ability when a new LiDAR is placed, which we will describe in Sec~\ref{sec31}.
To obtain the perceptual gain efficiently, we propose a perception predictor that learns to evaluate LiDAR placements using only a single frame point cloud data, which is detailed in Sec~\ref{sec32}.     

\begin{algorithm}
\caption{perceptual gain based greedy method}
\label{algorithm}
\KwIn{The number of LiDAR $M$, a set of feasible LiDAR positions $ \mathcal{P} = \{\bm{p}_1, \bm{p}_2, ..., \bm{p}_N\}$.
}
\KwOut{A set of selected LiDAR positions $\mathcal{S} = \{ \bm{p}_1^{*}, \bm{p}_2^{*}, ..., \bm{p}_M^{*} \}$. }
Initialize LiDAR position set $\mathcal{S} \leftarrow \varnothing$ \\
\For{$i \leftarrow 1$ \KwTo $M$}{
    Initialize selected LiDAR position $\bm{p}_i^* \leftarrow \varnothing$ \\
    Initialize maximum perceptual gain $g^* \leftarrow -\infty$ \\
    Calculate the perception score $k_{\mathcal{S}}$ given the selected LiDAR positions in $\mathcal{S}$
    \\
    \For{$\bm{p}$ in $\mathcal{P}$}{
        $\mathcal{S}^{\prime} \leftarrow \mathcal{S} \cup \{\bm{p}\} $ \\
        Calculate the perception score $k_{\mathcal{S}^{\prime}}$ for $\mathcal{S}^{\prime}$ \\
        Obtain the perceptual gain $g \leftarrow k_{\mathcal{S}^{\prime}} - k_{\mathcal{S}}$ \\    
        \If{$g > g^*$}{
            $\bm{p}_i^* \leftarrow \bm{p}$ \\
            $g^* \leftarrow g$
        }  
        $\mathcal{S}  \leftarrow \mathcal{S}^{\prime} \setminus \{\bm{p}\} $ 
    }
    $\mathcal{S} \leftarrow \mathcal{S} \cup \{\bm{p}_i^*\} $   
    \\
    $\mathcal{P} \leftarrow \mathcal{P} \setminus \{\bm{p}_i^*\}$
}
\textbf{return} $\mathcal{S}$
\end{algorithm}
\vspace{2mm}

\begin{figure*}[t]
   \centering
   \includegraphics[width=0.86\linewidth]{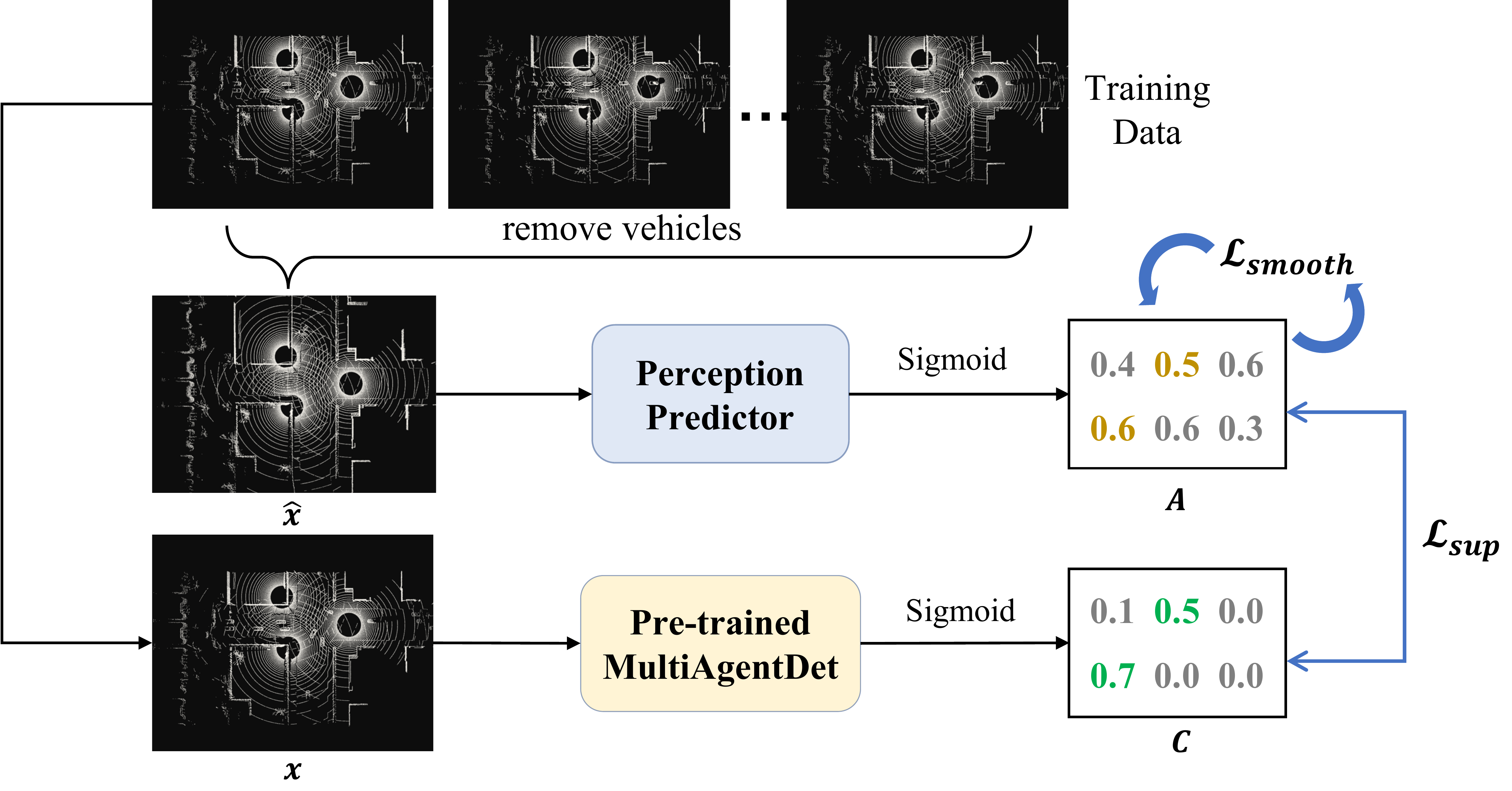}
   \caption{
        We obtain a frame of point cloud data without vehicle $\hat{\bm{x}}$ by selectively fusing different frames using labeled bounding boxes.
       Our perception predictor takes $\hat{\bm{x}}$ as input and predicts a perception ability map $\bm{A}$.
       We supervise $\bm{A}$ with $\mathcal{L}_{sup}$ using the confidence map $\bm{C}$ generated from a pre-trained multi-agent detection model.
       Note that we supervise $\bm{A}_{i,j}$ only if $\bm{C}_{i,j} > \text{threshold}$ (\ie, elements with color in the image above). 
       A smoothing loss $\mathcal{L}_{smooth}$ is also adopted for robustness.
   }
   \label{predictor}
\end{figure*}
\vspace{2mm}

\subsection{Perceptual gain based greedy method}

\label{sec31}
Given a feasible set of LiDAR positions $\mathcal{P} = \{\bm{p}_1, \bm{p}_2, ..., \bm{p}_N\}$, a straightforward way to optimize roadside LiDAR placement is to use a brute-force algorithm.
After evaluating all possible combinations of selected locations (\ie, $C^M_N$ decisions in total), the location with the highest average precision is selected. However, the running time of the brute-force algorithm becomes unacceptable as $N$ and $M$ become larger.
We observe that when we sequentially select the placement of the LiDAR, the obtained perceptual performance exhibits an intuitive diminishing returns property, also known as submodularity \cite{krause2014submodular,roberts2017submodular,huang2022optimizing}.
Inspired by this property, we propose a \textit{perceptual gain} based greedy algorithm to obtain an approximate optimal placement, as shown in the Algorithm \ref{algorithm}.

Our algorithm sequentially selects $M$ position from a set of feasible LiDAR positions $\mathcal{P}$.
In each loop, we tentatively place a LiDAR at each feasible position and compute the perceptual gain $g = k_{\mathcal{S}^{\prime}} - k_{\mathcal{S}}$ of the trial placement $\bm{p}$, where $k_{\mathcal{S}}$ and $k_{\mathcal{S}^{\prime}}$ are the perception scores before and after placing the LiDAR at $\bm{p}$ respectively.
The concept of perceptual gain denotes the increase in perceptual ability when an additional LiDAR is placed at $\bm{p}$.
The feasible position that yields the maximum perceptual gain $\bm{p}^*$ will be added to the selected LiDAR position set $\mathcal{S}$ and removed from $\mathcal{P}$.
After selecting $M$ LiDARs sequentially, we can obtain a set of optimized LiDAR positions $\mathcal{S}$.

Noted that our method does not need practical trial-and-error for selection. Instead, we only need to collect sequential point cloud data at each location once before using our method.
In the implementation, we select an ego LiDAR and then project point cloud data obtained by other LiDARs to the ego coordinate system, eliminating the need to place and remove LiDARs. This makes our method very efficient.

\subsection{Perception predictor for perceptual gain}
\label{sec32}
\textbf{Perception predictor.}
In our proposed greedy method for optimizing sensor placement, the calculation of the perceptual gain $g = k_{\mathcal{S}^{\prime}} - k_{\mathcal{S}}$ is the most important part. Nevertheless, it is time-consuming and computationally inefficient to evaluate the true detection performance for every possible position based on a whole dataset. To this end, we propose a \textit{perception predictor} that learns to predict the perception score for arbitrary sensor placement, as Figure \ref{predictor} shows.

In the training phase, we first randomly select multiple roadside LiDARs from $\mathcal{S}$ to form a subset $\hat{\mathcal{S}}$.
We then project the sequential training point cloud data collected from $\hat{\mathcal{S}}$ into the coordinate system of a randomly selected ego LiDAR.
Since we have sequential point cloud data labeled with vehicle bounding boxes, we can obtain a frame of point cloud data \textit{without} vehicle by selectively fusing different frames.
The selective fusion removes vehicles through bounding box annotations and obtains a frame of point cloud data termed as $\hat{\bm{x}}$. 
Our proposed perception predictor aims to predict a perception ability map $\bm{A} \in \mathbb{R}^{H \times W}$ for LiDAR placement $\hat{\mathcal{S}}$ given $\hat{\bm{x}}$, where
$\bm{A}_{i,j}$ denote model's perception ability for the placement $\hat{\mathcal{S}}$ if there is a vehicle at location $(i, j)$.

The motivation of the perception predictor is to learn the correlation between the perception ability and the pattern/distribution of the point cloud, enabling a quick evaluation of a roadside LiDAR placement.
We remove the vehicles in the point cloud data to retain the original distribution of the point cloud data.
The proposed perception predictor uses PointPillar~\cite{lang2019pointpillars} as the backbone to extract point cloud features from $\hat{\bm{x}}$ and then generate a perception ability map $\bm{A}$.
We supervise the perception predictor using the confidence map $\bm{C}$ produced by a pre-trained multi-agent detection model \cite{xu2022opv2v}.
Mathematically, the process of training is as follows:
\vspace{1mm}
\begin{equation}
\begin{gathered}
    \begin{aligned}
        \bm{A} &= \operatorname{Sigmoid}\left(\operatorname{predictor}(\hat{\bm{x}}) \right) \\
        \bm{C} &= \operatorname{Sigmoid} ( \operatorname{MultiAgentDet}(\bm{x}) ) \\
        \bm{K}_{i, j} &= 
        \begin{cases} 1, & \text { if } \bm{C}_{i, j} > \text{threshold} \\ 
        0, & \text { otherwise }\end{cases} 
        \end{aligned}
        \\
        \mathcal{L}_{sup} = \lVert \bm{K} \cdot (\bm{C} - \bm{A}) \rVert,
    \end{gathered}
\end{equation}
where $\bm{A}, \bm{C}, \bm{K} \in \mathbb{R}^{H \times W}$ denote the predicted confidence map, supervision confidence map, and supervision mask. $H$ and $W$ denote the output height and width.
$\mathcal{L}_{sup}$ denote the supervision loss.
Note that for different training point cloud data $x$, the vehicles appear in different locations.
We should only supervise $\bm{A}_{i,j}$ when vehicles appear to exist at the location $(i, j)$ of $x$ since $\bm{A}_{i,j}$ denote the perception ability if there is a vehicle at location $(i, j)$.
Thus, we set a threshold for the produced confidence map $\bm{C}$ and obtain the supervision mask $\bm{K}$, which enables the supervision on location $(i, j)$ only if $\bm{C}_{i,j} > \text{threshold}$.

\begin{table*}[t]
\setlength\tabcolsep{2.5pt}
\small
\centering
\caption{
    Quantitative experiments on the test set of the Roadside-Opt dataset. Three types of fusion methods are adopted to evaluate the average precision after selecting the LiDAR placement.
    The row ``Random'' indicates randomly selecting $M$ LiDAR positions from all the potential positions.
    The results of the Upper bound are obtained using the brute-force algorithm that tests every possible placement (\ie, $C^N_M$ in total).
}
\vspace{2mm}
\begin{tabular}{lccccccccc}
\hline
\multicolumn{1}{l|}{}            & \multicolumn{3}{c|}{Early Fusion}                       & \multicolumn{3}{c|}{Intermediate Fusion}                  & \multicolumn{3}{c}{Late Fusion}   \\ \cline{2-10}
\multicolumn{1}{l|}{}            & AP@0.3    & AP@0.5    & \multicolumn{1}{c|}{AP@0.7}     & AP@0.3     & AP@0.5     & \multicolumn{1}{c|}{AP@0.7}     & AP@0.3    & AP@0.5    & AP@0.7    \\ \hline
\multicolumn{10}{c}{ The number of LiDARs $M = 2$}                                                                                                                                                               \\ \hline
\multicolumn{1}{l|}{Random}      & $0.68\pm0.11$ & $0.64\pm0.10$ & \multicolumn{1}{c|}{$0.54\pm0.08$} & $0.48\pm0.08$  & $0.44\pm0.08$ & \multicolumn{1}{c|}{$0.31\pm0.04$}  & $0.40\pm0.18$ & $0.31\pm0.15$ & $0.22\pm0.09$ \\
\multicolumn{1}{l|}{Ours}        & $\textbf{0.82}$      & $\textbf{0.78}$      & \multicolumn{1}{c|}{$\textbf{0.56}$}       & $\textbf{0.74}$       & $\textbf{0.63}$       & \multicolumn{1}{c|}{$\textbf{0.40}$}       & $\textbf{0.65}$      & $\textbf{0.56} $     & $\textbf{0.31}$      \\ \hline
\multicolumn{1}{l|}{Upper bound} &   $0.85$    &    $0.82$   & \multicolumn{1}{c|}{$0.72$}       &   $0.78$    &    $0.71$    & \multicolumn{1}{c|}{$0.52$}     &    $0.70$    &    $0.61$   &  $0.43$        \\ \hline
\multicolumn{10}{c}{The number of LiDARs $M = 3$}                                                                                                                                                               \\ \hline
\multicolumn{1}{l|}{Random}      & $0.75\pm0.20$ & $0.73\pm0.19$ & \multicolumn{1}{c|}{$0.62\pm0.18$}  & $0.57\pm0.10$  & $0.54\pm0.08$ & \multicolumn{1}{c|}{$0.43\pm0.07$}  & $0.49\pm0.13$ & $0.40\pm0.12$ & $0.33\pm0.10$ \\
\multicolumn{1}{l|}{Ours}        & $\textbf{0.89}$      & $\textbf{0.88}$      & \multicolumn{1}{c|}{$\textbf{0.81}$}       & $\textbf{0.80}$       & $\textbf{0.75}$       & \multicolumn{1}{c|}{$\textbf{0.56}$}       & $\textbf{0.71} $     & $\textbf{0.65}  $    & $\textbf{0.45}$      \\ \hline
\multicolumn{10}{c}{The number of LiDARs $M = 4$}                                                                                                                                                               \\ \hline
\multicolumn{1}{l|}{Random}      &$ 0.79\pm0.17$ & $0.77\pm0.16$ & \multicolumn{1}{c|}{$0.70\pm0.13$}  & $0.70\pm0.14$ & $0.69\pm0.14$  & \multicolumn{1}{c|}{$0.61\pm0.12$} & $0.65\pm0.16$ &$ 0.61\pm0.15$ &$ 0.51\pm0.12$ \\
\multicolumn{1}{l|}{Ours}        & $\textbf{0.92}$      &$ \textbf{0.91} $     & \multicolumn{1}{c|}{$\textbf{0.89}$}       & $\textbf{0.83}$       & $\textbf{0.81}$       & \multicolumn{1}{c|}{$\textbf{0.65}$}       & $\textbf{0.78}$      &$\textbf{ 0.73} $     &$ \textbf{0.56}  $    \\ \hline
\end{tabular}
\vspace{8mm}
\label{tab1}
\end{table*}

To increase the robustness and generalization, we also add a smoothing loss $\mathcal{L}_{smooth}$ to the overall loss $\mathcal{L}$:
\begin{equation}
\begin{gathered}
    \mathcal{L} = \mathcal{L}_{sup} + \gamma \mathcal{L}_{smooth} \\
    \begin{aligned}
    \mathcal{L}_{smooth} &= \frac{1}{HW} \sum_{i,j} 
\left|\bm{A}_{i,j} - \bm{A}_{i-1,j}\right| + \left|\bm{A}_{i,j} - \bm{A}_{i,j-1}\right| \\
    &+ \left|\bm{A}_{i,j} - \bm{A}_{i+1,j}\right| + \left|\bm{A}_{i,j} - \bm{A}_{i,j+1}\right|.
    \end{aligned}
\end{gathered}
\end{equation} 
Intuitively, the perceptual abilities of neighboring regions do not vary much.
So for each element in $\bm{A}$ we compute the difference between it and its neighbors.
We then include $\mathcal{L}_{smooth}$ in the overall supervision $\mathcal{L}$ with a relatively small factor $\gamma$.

\textbf{Perceptual gain.}
The trained perception predictor learns to find the correlation between the perception ability and the pattern/distribution of the point cloud, which allows a quick evaluation of a roadside LiDAR placement.
Thus, we accumulate the perception ability map $\bm{A}$, which we refer to as the perception score $k$:
\begin{equation}
    k = \sum_{i,j} \bm{A}_{i,j}.
\end{equation}
Through tentatively placing a LiDAR at every feasible position, we compute the perceptual gains $g = k_{\mathcal{S}^{\prime}} - k_{\mathcal{S}}$ of the trial placement $\mathcal{S}^{\prime}$, where $k_{\mathcal{S}}$ and $k_{\mathcal{S}^{\prime}}$ are the perception scores before and after placing the LiDAR respectively.
The feasible position that maximizes perceptual gain will be added to the selected LiDAR position set $\mathcal{S}$ and removed from $\mathcal{P}$.

\begin{figure*}[t]
  \centering
  \includegraphics[width=1.0\linewidth]{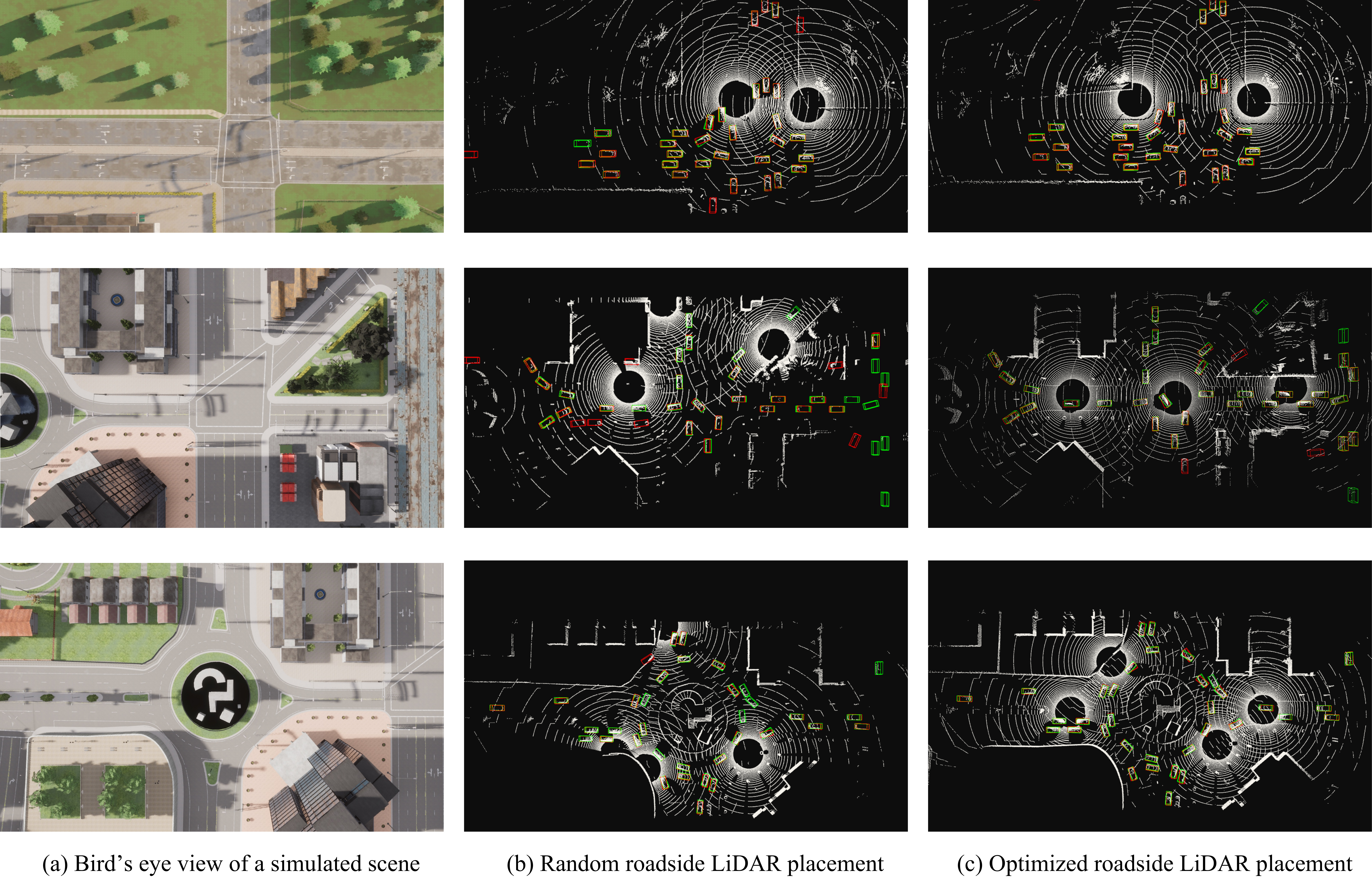}
  \caption{
    We visualize one frame from each of the three test scenes, where $2, 3, 4$ LiDARs are placed in each scene.
    (a) are the simulated scenes of the test set.
    (b) are the visualization of the baseline, which generate random placements from all the potential locations.
    (c) are the visualization of optimizing the placement using our proposed method.
    The \textcolor{green}{green} and \textcolor{red}{red} bounding boxes denote the ground truth vehicles and the detected vehicles using the pre-trained detection model, respectively.
  }
  \label{opt_vis}
\end{figure*}
\vspace{2mm}

\section{Experiment}
\subsection{Dataset and settings}
\textbf{Dataset.}
In recent years, an increasing number of autonomous driving datasets have been released.
In addition to single-agent datasets such as KITTI \cite{geiger2013vision} and nuScenes \cite{caesar2020nuscenes}, there are several multi-agent/V2X datasets that contain roadside LiDAR data, including the synthetic dataset OPV2V \cite{xu2022opv2v}, V2XSet \cite{xu2022v2x}, V2XSim \cite{li2021learning}, and realistic dataset DAIR-V2X \cite{yu2022dair}, Rope3D \cite{ye2022rope3d}.
However, all these multi-agent datasets contain only 1 or 2 roadside LiDARs in each scene, which is far from what we need for the experiment.
Since there is no suitable dataset for roadside LiDAR optimization research, we build our own dataset named Roadside-Opt using the realistic CARLA simulator \cite{dosovitskiy2017carla} and simulation tool OpenCDA \cite{xu2021opencda}.
With the simulator, we build our dataset so that we can fairly evaluate different LiDAR placements with all other environmental factors fixed, such as the trajectories of the ego vehicle and surrounding objects.

To collect the Roadside-Opt dataset, we build 10 different scenes in CARLA Town $3, 4, 5, 6$ with recorded dense traffic flow.
For each scene, we set up at least $15$ feasible roadside LiDAR positions and spawn at least $50$ vehicles in the scene.
Feasible roadside LiDAR positions are collected  by finding streetlights and traffic lights on the scene.
If there are only a few streetlights or traffic lights, we randomly generate multiple feasible roadside LiDAR positions in the scene.
The collected Roadside-Opt dataset contains $37,641$ point cloud frames from $126$ roadside LiDARs in total.
Each roadside LiDAR has a rotation frequency of $20$Hz and approximately $20000$ points per frame.
We take $40\%$ of the scenes as the test set and the remaining data as the training set for all the experiments.

\textbf{Implementation details.}
For every experiment, we customize the range of each LiDAR to be $l=140 \text{m}$, $w=40 \text{m}$, $h=4 \text{m}$ following the settings of \cite{xu2022opv2v,xu2022v2x}.
We pre-train the multi-agent 3D object detection method \cite{xu2022opv2v} in the training set of Roadside-Opt for the supervision of perception predictor.
The factor $\gamma$ for the smoothing loss $\mathcal{L}_{smooth}$ is set to $0.1$.
The threshold set for calculating $\mathcal{L}_{sup}$ for training the perception predictor is $0.2$.
 
We follow the existing multi-agent perception papers~\cite{xu2022opv2v,xu2022v2x} to use Average Precision (AP) as the metric to evaluate the perception performance after selecting a roadside LiDAR placement.
We compute APs through three types of fusion methods including 
early fusion, intermediate fusion, and late fusion.
Early fusion directly projects and aggregates raw LiDAR point clouds from other agents.
Late fusion gathers all detected outputs from all agents and applies Non-Maximum Suppression (NMS) to obtain the final results.
For intermediate fusion, we evaluate APs using a well-known multi-agent 3D detection method \cite{xu2022opv2v}.
For a fair comparison, all three types of fusion methods use PointPillar~\cite{lang2019pointpillars} as the backbone and are trained on the training set of the Roadside-Opt dataset for $50$ epochs.
We use the Adam optimizer~\cite{kingma2014adam} with an initial learning rate of 0.001 for all the models.
Experiments compared with \cite{hu2022investigating} and a coverage/density baseline that first maximizes the coverage of LiDARs and then the density of the point cloud can be found in the supplemental material.

\begin{figure}[t]
  \centering
  \includegraphics[width=1.0\linewidth]{imgs/opt_num_long.pdf}
  \caption{
     Visualization of the process of selecting optimal locations that maximize the perceptual gain sequentially (from $1$ LiDAR to $5$ LiDARs).
  }
  \label{opt_num}
\end{figure}

\subsection{Experimental Results}

\textbf{Quantitative experiments.}
We conduct quantitative experiments on the test set of our proposed Roadside-Opt dataset and set the number of LiDARs $M = 2,3,4$, as shown in Table~\ref{tab1}.
We randomly select a combination of LiDAR positions for $3$ times and calculate the mean and variance of the obtained APs as a baseline.
We employ three types of multi-agent fusion strategies, including early fusion, intermediate fusion, and late fusion.
The perceptual gain based greedy method sequentially computes the feasible position that can maximize the perceptual gain.
And our proposed perception predictor can efficiently evaluate 
perceptual gain for a trial location.
By incorporating the perceptual gain based greedy method and the perception predictor, our method significantly outperforms the random baseline method.
We also compute the upper bound of perceptual performance, which is to use the brute-force algorithm that tests every possible placement (\ie, $C^N_M$ in total).
Since the brute-force algorithm is very time-consuming, we only experiment on the cases where $M=2$. We can observe that the result obtained by our greedy method is close to the upper bound, which shows that our method is able to obtain approximately optimal solutions and also dramatically reduce the computation time.
The calculation time of the brute-force method (upper bound) exceeds $11$ hours and $46$ hours when the number of LiDAR is 2 and 3, as it requires evaluating the Average Precision (AP) on the test set for every possible combination.
While our proposed greedy method calculates the optimal placement within $10$ seconds.

\begin{table}[t]
\caption{
    Quantitative experiments on the process of selecting optimal locations that sequentially maximize perceptual gain (from $1$ LiDAR to $5$ LiDARs). 
} 
\centering
\label{tab2}

\begin{tabular}{lccc}
\hline
                          & \multicolumn{3}{c}{Intermediate Fusion} \\ \hline
\multicolumn{1}{l|}{}     & AP@0.3      & AP@0.5      & AP@0.7      \\ \hline
\multicolumn{4}{c}{ The number of LiDARs $M = 1$}                                           \\ \hline
\multicolumn{1}{l|}{Ours} &     $0.50$        &    $0.38$       &     $0.17$               \\ \hline
\multicolumn{4}{c}{ The number of LiDARs $M = 2$}                                           \\ \hline
\multicolumn{1}{l|}{Ours} &   $0.81$    &      $0.69$    &   $0.38$              \\ \hline
\multicolumn{4}{c}{ The number of LiDARs $M = 3$}                                           \\ \hline
\multicolumn{1}{l|}{Ours} &  $0.87$    &   $0.81$   &    $0.54$         \\ \hline
\multicolumn{4}{c}{ The number of LiDARs $M = 4$}                                           \\ \hline
\multicolumn{1}{l|}{Ours} &  $0.90$    &  $0.88$    &   $0.67$           \\ \hline
\multicolumn{4}{c}{ The number of LiDARs $M = 5$}                                           \\ \hline
\multicolumn{1}{l|}{Ours}  &  $0.91$    &   $0.90$   &    $0.76 $         \\ \hline
\end{tabular}
\end{table}
\vspace{3mm}

\textbf{Visualization of optimized example.}
We perform the experiment on the test set of the Roadside-Opt dataset and visualize some of the examples, as shown in Figure~\ref{opt_vis}.
We visualize one frame from each of the three test scenes, where $2, 3, 4$ LiDARs are placed in each scene.
Column (a) shows the bird's eye view of three simulated scenes in the CARLA simulator.
Column (b) shows the baseline of randomly select $M$ LiDARs from all the feasible setup locations. 
Column (c) indicates the visualization of optimizing
the placement with our proposed method.
We can observe that by optimizing the placement, the selected LiDAR positions are more reasonable, which covers more areas in the scene.
The red box obtained by optimal LiDAR placement overlaps more with the green box, indicating an improvement in perceptual performance.

\subsection{Analysis}

\textbf{Visualization of the sequential selection.}
In the method section, we propose to use a perceptual gain based greedy method to sequentially select optimal LiDAR based on the calculated perceptual gain.
In Figure~\ref{opt_num}, we visualize the process of sequentially selecting optimal locations, where the number of LiDAR increase from $1$ to $5$.
All experiments use the same pre-trained multi-agent 3D detection model for a fair comparison.
The left column visualizes different roadside LiDAR placements under the same point cloud data from the test set.
In the greedy method, we sequentially select the locations to deploy the LiDAR, which covers more and more areas of the intersection and produces increasingly accurate red bounding boxes, as shown in the left column.
We also visualize the predicted perceptual map for each step, as shown in the right column.
As the number of LiDAR deployed increases, the perception ability map expands to cover most of the intersection.

\textbf{Quantitative analysis of selecting optimal LiDAR sequentially.}
To quantitatively analyze the process of the greedy algorithm, we evaluate the average precision for the $5$ placement shown in Table~\ref{tab2} using the intermediate fusion method~\cite{xu2022opv2v}.
As the number of placed LiDARs increases, the obtained perceptual performance rapidly increases as we select the location that can maximize the perceptual gain.
The perceptual performance also exhibits an intuitive diminishing returns property as we place more LiDARs.

\begin{figure}[t]
  \centering
  \includegraphics[width=1.0\linewidth]{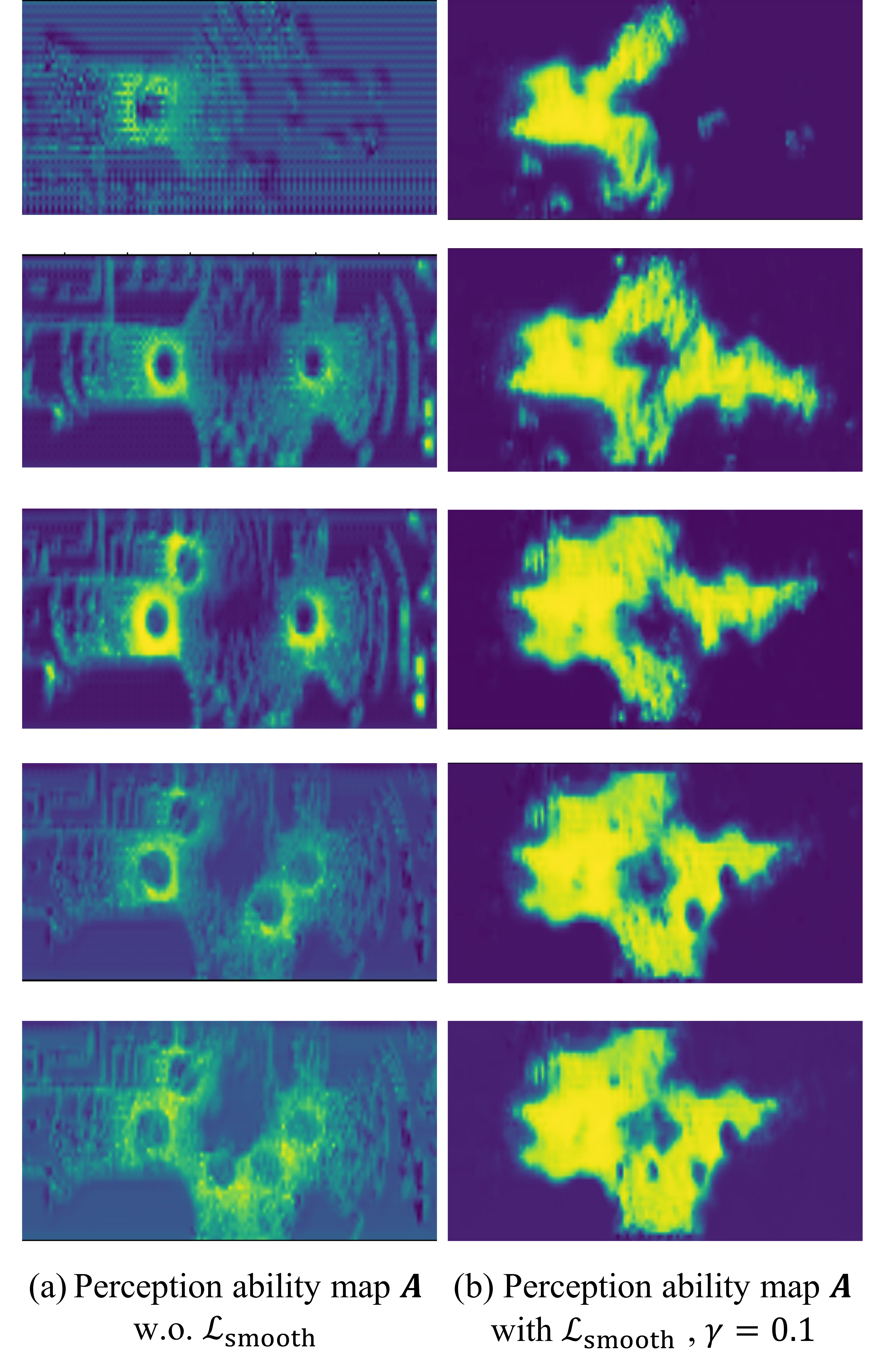}
  \caption{
     Visualization of the perception ability map $\bm{A}$ in the process of selecting optimal locations sequentially using our method. The detail of the scene is shown in Figure~\ref{opt_num}.
     $\bm{A}$ obtained by perception predictor trained with and without the smoothing loss $\mathcal{L}_{smooth}$ are shown in column (b) and (a) respectively.
  }
  \label{smooth}
\end{figure}
\vspace{2mm}

\textbf{Visualization of perception ability map.}
In the method section, we propose to use the perception predictor to quickly evaluate the LiDAR placement.
We train the perception predictor using $\mathcal{L}_{sup}$, which uses the confidence map produced by the off-the-shelf detection model for supervision.
A smoothing loss $\mathcal{L}_{smooth}$ is also adopted with a factor $\gamma$ to increase the robustness and generalization.
We visualize the perception ability map $\bm{A}$ obtained from the perception predictor trained with and without smoothing loss, as shown in Figure~\ref{smooth}.
We can observe that the perception ability maps in the left column are sparse and discontinuous, while the perception ability maps in the right column are more continuous and smoother.
This is because when training the perception predictor, the vehicles do not appear in every location of the training frames.
Thus the confidence map $\bm{C}$ used for supervision is sparse since we set a threshold to obtain the supervision mask $\bm{K}$.
Intuitively, the perception abilities of neighboring regions do not vary much.
We then add a smoothing loss for the training of the perception predictor with an appropriate factor $\gamma$, which makes the obtained perception ability map more continuous.

\textbf{Limitations.}
The main limitation of our method and also the current methods is the application to the real world.
Both ours and existing papers \cite{cai2022analyzing,hu2022investigating,ma2021perception} that study the LiDAR configuration problem are all tested in the simulation environment, as evaluating placements with all other environmental factors fixed in the real world is infeasible.
To adapt our method to the real world, users need to build digital towns in CARLA simulator that mimic real-world road topology and spawn realistic traffic flow in the simulator.
For instance, OPV2V \cite{xu2022opv2v} has created a digital town of Culver City, Los Angeles, following the aforementioned steps, which accurately replicates the real-world environment.

\section{Conclusion}
In this paper, we aim to optimize the placement of roadside LiDARs for multi-agent cooperative perception in autonomous driving, which is a crucial but rarely studied problem.
We propose a novel perceptual gain based greedy method that selects optimized positions sequentially.
For each step in the sequence, we greedily choose the position that can maximize the perceptual gain.
Perceptual gain indicates the increase in perceptual ability when a new LiDAR is placed.
To obtain the perceptual gain, we design a perception predictor that learns to evaluate a LiDAR placement using a single frame of point cloud data without vehicles.
Since there is no suitable dataset for the experiment, we build our own Roadside-Opt dataset containing scenes with multiple roadside LiDARs.
Extensive experiments are performed on the newly proposed dataset, demonstrating the effectiveness of our method.
Our proposed greedy method can obtain approximately optimal solutions and also greatly reduce the computation time.
We believe our method can also be used in other tasks that require optimization on LiDAR placement.

\textbf{Acknowledgement.}
This work was supported  in part by Shanghai Artificial Intelligence Laboratory and National Key R\&D Program of China (Grant No. 2022ZD0160104) and the Science and Technology Commission of Shanghai Municipality (Grant No. 22DZ1100102), in part by the National Key R\&D Program of China under Grant 2022ZD0115502, in part by the National Natural Science Foundation of China under Grant 62122010, in part by the Fundamental Research Funds for the Central Universities, in part by Tier 2 grant MOE-T2EP20120-0011 from the Singapore Ministry of Education.

{\small
\bibliographystyle{ieee_fullname}
\bibliography{egbib}
}

\end{document}